# Automatic Conditional Generation of Personalized Social Media Short Texts

Ziwen Wang[1,2], Jie Wang[3], Haiqian Gu[2,4], Fei Su[2,4] and Bojin Zhuang[3]

[1]School of Science, Beijing University of Posts and Telecommunications, China
[2] Beijing Key Laboratory of Network System and Network Culture
[3] Ping An Technology (Shenzhen) Co., Ltd.
[4]School of Information and Communication Engineering, Beijing University of Posts and Telecommunications, China
wangziwen@bupt.edu.cn, wangjie388@pingan.com.cn, mixiu@bupt.edu.cn, sufei@bupt.edu.cn, zhuangbojin232@pingan.com.cn

**Abstract.** Automatic text generation has received much attention owing to rapid development of deep neural networks. In general, text generation systems based on statistical language model will not consider anthropomorphic characteristics, which results in machine-like generated texts. To fill the gap, we propose a conditional language generation model with Big Five Personality (BFP) feature vectors as input context, which writes human-like short texts. The short text generator consists of a layer of long short memory network (LSTM), where a BFP feature vector is concatenated as one part of input for each cell. To enable supervised training generation model, a text classification model based convolution neural network (CNN) has been used to prepare BFP-tagged Chinese micro-blog corpora. Validated by a BFP linguistic computational model, our generated Chinese short texts exhibit discriminative personality styles, which are also syntactically correct and semantically smooth with appropriate emoticons. With combination of natural language generation with psychological linguistics, our proposed BFP-dependent text generation model can be widely used for individualization in machine translation, image caption, dialogue generation and so on.

## 1 Introduction

Natural language generation (NLG) has been intensively studied owing to its important applications such as automatic dialogue generation [1], machine translation [2], text summarization [3], image captions [4] and so on. Based on deep learning approaches, much research effort has been paid to improving the quality of automatically-generated texts in terms of syntax and semantics. For instance, Karpathy (2015) put forward a character-level recurrent neural network (Char-RNN) for generation tasks [5]. With contextual controlling conditions, sequence to sequence (Seq2Seq) models for machine translation and smart dialogue generation were first introduced by Google Brain in 2014 [6]. Furthermore, Guo (2015) proposed reinforcement learning (RL) scheme which built a deep Q-network as a language generator [7]. Zhang et al (2016) tried generative adversarial networks (GAN) theory for NLG which employed CNN as the discriminator and LSTM as the generator [8].

Most newly-developed language generation models concentrate on grammatical correction and semantic correlation, however, an essential issue has not been considered: in comparison with texts composed by human beings, machine-generated ones lack of personalities. Psychological linguistics demonstrates that contents and the way people write show their personalities, which could hardly be achieved by NLG robots. Nevertheless, language generation with personalities is demanding in some important application cases, including intelligent customer service and chat bots.

To fill this gap, Li et al (2016) encoded users' background information in distributed embeddings for neural response generation [9]. On contrast, we encode users' BFP information as contextual control in natural language generation. In this paper, we propose an efficient BFP computation model through classifying social media texts based on a CNN network. By concatenating BFP feature vector as partial input of LSTM cell, a conditional text generation model has been trained which write personality-discriminative short texts.

The rest of this paper is organized as follows. Section 2 presents our conditional language generation model, consisting of a CNN-based BFP classification sub-model and an LSTM-based text generator. Section 3 shows and analyzes generation results and Section 4 discusses the conclusions and our limitations.

## 2 The Conditional Language Generation Model

The BFP theory is one of the most important psychological theories for profiling personality of human beings. Amongst it, five big factors have been defined individually as extraversion (E), agreeableness (A), conscientiousness (C), neuroticism (N) and openness (O). As demonstrated by previous reports [10] [11] [12], people's BFP traits can be inferred by texts they shared on social media. Based on Linguistic Inquiry and Word Count (LIWC), the BFP of social media users can be analyzed by counting word frequency of certain psychological lexicons. Here, we combine natural language generation with psychological linguistics to propose a BFP-dependent short text generation model. The BFP-based generation model's architecture is shown in Fig. 1. A CNN-based text classification model is trained to achieve BFP representations of micro-blog texts and provide tagged corpora for language generation model training. The main text generator is based on a LSTM language model, which receives a BFP feature vector as the contextual input. With preprocessed Chinese corpora, the generation model is supervise-trained and then used for composing text sequences automatically. Eventually, BFP properties of such generated short texts have been validated based on the psychological linguistic computation method.

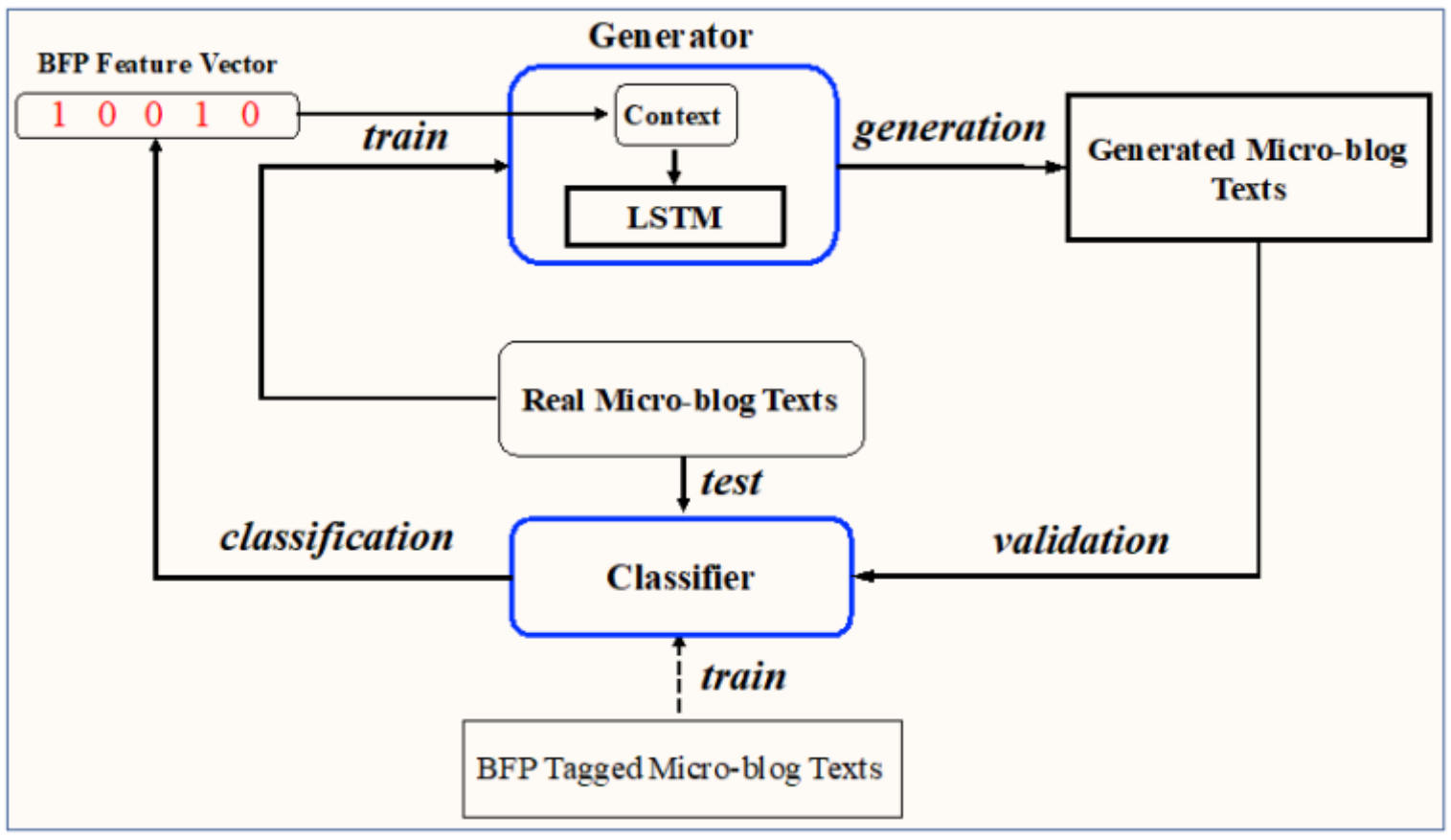

**Fig. 1.** The architecture of the BFP-dependent conditional generation model.

## 2.1 Psychological Text Classification Model

Lin et al identified the association between Chinese words and phrases and BFP traits based on SC-LIWC (a simplified Chinese version of LIWC developed by Gao et al. [13]) [12]. Based on SC-LIWC and word-frequency-to-BFP mapping matrix, we have prepared BFP-labelled corpora with the collected social media texts of 17,000 Sina Weibo users (http://www.weibo.com) to train a CNN-based Psychological text classification model. After being supervise-trained, the text classification model can be used to collect social media texts into bins of high or low BFP traits.

The text classifier is designed as shown in Fig. 2. Every input Chinese word is embedded into k dimensional representation vector by word2vec [14]. A convolution operation is then applied with a filter w $\in \mathbb{R}^{mk}$, which acts on a window of $m$ words to abstract a higher level feature, in practice, $m$ is set as 3 and Relu activation function is applied after the convolution operation. To capture the most important linguistic features, a max-over-time pooling operation is applied in each feature map. Eventually, all captured features are passed to a fully connected layer with SoftMax output to compute the probability distribution over the five big factors (i.e. E, A, C, N, O).

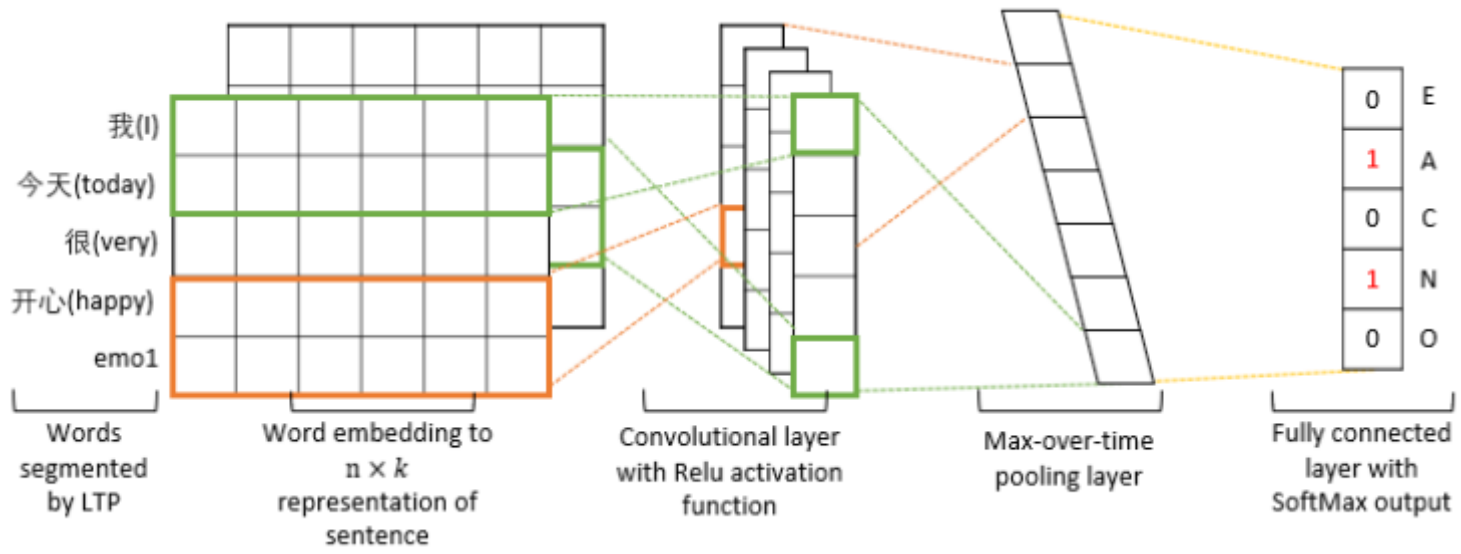

**Fig. 2.** The network architecture of the text classifier of BFP.

The prepared BFP-labelled Chinese corpora was randomly shuffled and then split into train and validation datasets with ratio of 9:1 for cross-validation of the model parameters. The classification accuracy in each BFP dimension is shown in Table 1. Such a high accuracy demonstrates the reliability of this text classifier, which will be used to label larger social media text corpora for training language generation model.

**Table 1.** The prediction accuracies of validation datasets of the classification model.

| Dimension | Accuracy |
|---|---|
| Extraversion | 0.9608 |
| Agreeableness | 0.9508 |
| Conscientiousness | 0.9713 |
| Neuroticism | 0.9806 |
| Openness | 0.9640 |

## 2.2 The Text Generator

The network architecture of text generator, as shown in Fig. 3, compromises an unrolled LSTM sequence. A word-to-vector embedding layer is firstly used to convert input Chinese words into dense tensors. Parameters of embedding layer will be fine-tuned along with training of the generation model.

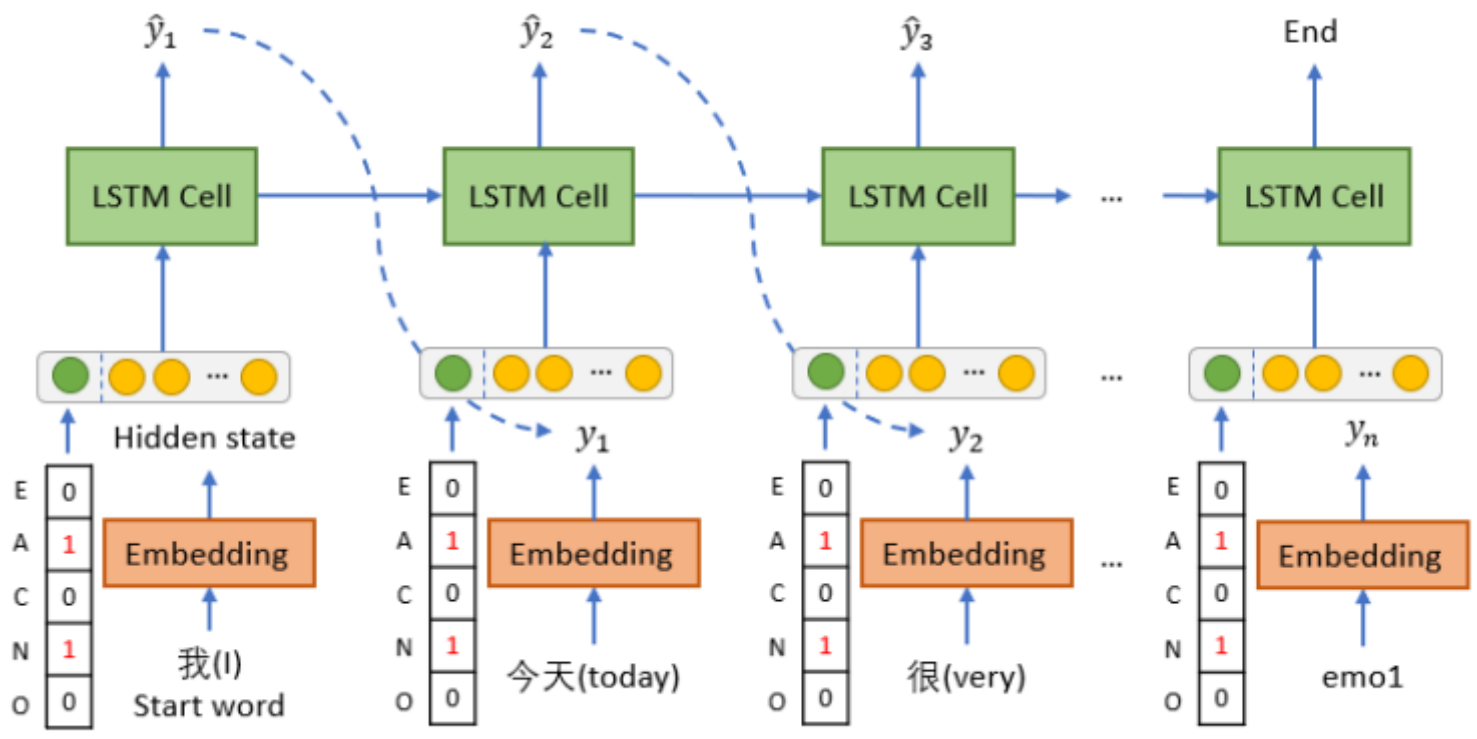

**Fig. 3.** The network architecture of the text generator.

The output word at time $y_t$ is predicted by the LSTM cell memory and previous output word $y_{t-1}$ combined with a global contextual BFP vector of five dimensions, where 1 or 0 is used to represent high or low BFP polarities. The output state of LSTM cell is multiplied by a fully connected layer and eventually a SoftMax layer to determine the occurrence probability distribution of each word in vocabulary. In the training process, the cross-entropy loss between output word sequences and the targets is used for model optimization.

During the process of decoding, a picked first word or sentence should be input to start sequence generation. In practice, a pool of random words was used to select the seed word to enhance the generation variety. The generation loop will continue until the predefined maximum length is reached, or be forced to stop when repeating words or phrases are detected.

# 3 Text Generation Results

Short texts of 6403 Sina Weibo users were selected as training corpora for our conditional text generation model. In order to guarantee the BFP discriminability of training corpora, invalid user IDs including advertisement accounts were cleaned according to predefined filtering criteria. After classified by the above CNN-based text classification model, preprocessed short texts and corresponding BFP representations were then prepared to train the conditional text generator.

## 3.1 Newly-Generated Text Samples

Text samples from our language generation model under condition of Extraversion dimension are shown in Table 3. The appropriate emoticons have been included to enhance the vividness of generated texts.

According to the below table, it can be observed that texts generated under condition of low extraversion are more objective and philosophical and does not exhibit too intense feelings with less emoticons. Oppositely, texts generated under condition of high extraversion are with strong feeling and more emoticons. These results are consistent with our common understanding of human personality. Introverts often like thinking deeply while outgoing people like to pour out their emotions. In addition, generated texts are fluent and logically correlated with proper emotions, which will be demonstrated in Section 3.3.

**Table 3.** Generated text samples under condition of Extraversion dimension.

| Low Extraversion case | High Extraversion case |
|---|---|
| 世界上最难的东西，是一定要自己承担的，也会有结果。 | 感谢我们的工作人员辛苦了 |
| 每个人都有自己的痛苦，也许你觉得一个人生活没有什么，但人生如此艰难。 | 听了好几遍 我的宝贝是不是都好可爱啊！ |
| 每个人都有自己的委屈，所以，我们才能 | 谁的人生，总是在令人心疼 |

对自己善良。

### 3.2 The Human Evaluation of Generated Texts

Since NLG for novel text generation is totally different from machine translation, where BLEU, METEOR and other indicators can be used for objective evaluation, human judgement is employed here to evaluate newly-generated short texts. Similar to the previous work [9], 30 real micro-blog texts and 30 model-generated ones from conditional and unconditional models are mixed and scored by 50 volunteers. Once regarded as human composition, it gets +1 human-like score, otherwise 0. At the meantime, volunteers are asked to mark coherence scores (1-5) of all texts, according to their fluency and logicality. The average scores of model-generated texts and real ones are calculated and compared in Table 4.

**Table 4.** The average scores of model-generated and real texts.

| Method | Human-like score | Coherence score |
|---|---|---|
| BFP-conditioned generation | **0.5387** | **3.972** |
| Unconditional generation | 0.4827 | 3.536 |
| The real texts | 0.7360 | 4.201 |

As shown in the above table, human-like and coherence scores of texts from BFP conditional generation model is comparable to real ones, which indicates that our generation system performs comparably to real human compositions in terms of fluency and logicality. As a consequence, our proposed BFP-based psychologically conditional generation model can applied to other NLG missions.

### 3.3 The BFP Scores of Generated Texts

With different BFP input conditions, 50000 short texts are totally generated. The BFP scores of model-generated texts are computed by SC-LIWC psychological analysis method which has been demonstrated effective. To verify the personalized effects of our proposed conditional generation model, we calculated the percentage of newly-generated texts with BFP scores located in three levels (i.e. low, medium, and high), which were determined by using the same standard mentioned in Section 2.1. Moreover, the corresponding percentage of machine-composed texts from unconditional generation model is also shown as a reference in Table 5.

**Table 5.** The percentage of newly-generated texts with BFP scores located in three levels.

| Dimension | Condition | The newly-generated texts with BFP scores located in | | |
|---|---|---|---|---|
| | | Low level | Medium level | High lever |
| Extraversion | Low condition | 90.91% | 3.50% | 5.59% |
| | High condition | 4.30% | 19.35% | 80.65% |
| | Unconditional | 35.67% | 32.48% | 31.85% |
| Agreeableness | Low condition | 58.76% | 41.24% | 0.00% |
| | High condition | 11.24% | 37.08% | 51.68% |
| | Unconditional | 18.47% | 72.61% | 8.92% |
| Conscientiousness | Low condition | 67.16% | 26.87% | 5.97% |
| | High condition | 4.79% | 31.51% | 63.70% |
| | Unconditional | 24.21% | 61.78% | 14.01% |
| Neuroticism | Low condition | 66.02% | 33.98% | 0.04% |
| | High condition | 1.54% | 17.69% | 80.77% |
| | Unconditional | 10.83% | 45.22% | 43.95% |
| Openness | Low condition | 75.56% | 23.33% | 1.11% |
| | High condition | 1.00% | 18.00% | 81.00% |
| | Unconditional | 25.48% | 26.75% | 47.77% |

As observed in Table 5, the distribution of BFP scores of newly-generated texts from conditional generation model are consistent with their corresponding BFP input conditions, while the BFP scores of newly-composed texts from unconditional generation model almost distribute normally. Here, we define generation accuracy as consistency between computed BFP score level and input BFP condition of newly-generated texts. As a result, the average generation accuracy in each dimension of our proposed conditional generation model is 71.62%, which could be difficultly achieved by traditional unconditionally NLG methods.

## 4 Conclusions and Limitations

In this paper, we have proposed and demonstrated a personalized short text generation model, which consists of a CNN-based text classifier and a conditional LSTM-based text generator. Among them, the CNN-based text classifier has been trained in supervise mode to achieve BFP polarity of social media short texts. With BFP feature vectors as contextual input, a conditional LSTM-based text generator has been trained to compose short texts with discriminative personality styles. Additionally, human evaluation has been conducted to prove the fluency and logicality of the generated texts. Moreover, BFP traits of newly-generated texts have computed based on the SC-LIWC BFP prediction method. High BFP generation accuracy of our conditional generation model facilitates psychologically controllable NLG in some promising industrial applications of artificial intelligent. For example, human-like chat-bots can be benefited from our technique, which can generate personality-specific dialogues according to users' background profiles built with the BFP prediction model.

However, there still exist several limitations in our study. For instance, our generation model based on LSTM is suitable for short text generation such as micro-blog texts, but hardly composes semantically-correlated long sentences and paragraphs. Additionally, in case of instant dialogue systems, mood states of users should be taken account of for automatic dialogue generation in terms of empathy effects.

## Acknowledge

This work is supported by Chinese National Natural Science Foundation (61471049, 61532018).